%% file: emnlp2017.tex
\documentclass[11pt,letterpaper]{article}
\usepackage{emnlp2017}
\usepackage{times}
\usepackage{latexsym}

\usepackage{url}
\usepackage{verbatim}
\usepackage{color}
\usepackage{lipsum}
\usepackage{subfig}
\usepackage{graphicx}
\usepackage{tabularx}
\usepackage{paralist}
\usepackage{comment}
\usepackage{xcolor}

\emnlpfinalcopy

\def\emnlppaperid{***}

\newcommand\BibTeX{B{\sc ib}\TeX}

\newcommand{\TG}{\textit{target word}}

\title{All that is English may be Hindi: Enhancing language identification through automatic ranking of likeliness of word borrowing in social media}

\author{$^1$Jasabanta Patro, $^2$Bidisha Samanta, $^2$Saurabh Singh, \\\textbf{ $^2$Abhipsa Basu,$^2$Prithwish Mukherjee, $^3$Monojit Choudhury, $^4$Animesh Mukherjee}\\
$^{1,2,4}$ Indian Institute of Technology Kharagpur, India -- 721302\\
$^6$Microsoft Research India, Bangalore -- 560001 \\ 
$^1$jasabantapatro@iitkgp.ac.in, $^3$monojitc@microsoft.com, $^4$animeshm@cse.iitkgp.ernet.in \\
}

\date{}

\begin{document}

\maketitle

\input{00Abstract}
\input{10Introduction}

\input{20Relatedwork}
\input{30Methodology}

\input{40Experiments}
\input{50Results}

\input{60Conclusion}

\bibliography{emnlp2017}
\bibliographystyle{emnlp_natbib}

\end{document}

%% file: 00Abstract.tex
\begin{abstract}

In this paper, we present a set of computational methods to identify the likeliness of a word being borrowed, based on the signals from social media. In terms of Spearman's correlation values, our methods perform more than two times better ($\sim 0.62$) in predicting the borrowing likeliness compared to the best performing baseline ($\sim 0.26$) reported in literature. Based on this likeliness estimate we asked annotators to re-annotate the language tags of foreign words in predominantly native contexts. In 88\% of cases the annotators felt that the foreign language tag should be replaced by native language tag, thus indicating a huge scope for improvement of automatic language identification systems.

\end{abstract}

    

%% file: 10Introduction.tex
\section{Introduction}
In social media communication, multilingual people often switch between languages, a phenomenon known as {\em code-switching} or {\em code-mixing} ~\cite{auer}. This makes \textit{language identification and tagging}, which is perhaps a pre-requisite for almost all other language processing tasks that follow, a challenging problem~\cite{barman2014code}. In code-mixing people are subconsciously aware of the foreign origin of the code-mixed word or the phrase. A related but linguistically and cognitively distinct phenomenon is {\em lexical borrowing} (or simply, {\em borrowing}), where a word or phrase from a foreign language say $L_2$ is used as a part of the vocabulary of native language say $L_1$. For instance, in Dutch, the English word ``sale'' is now used more frequently than the Dutch equivalent ``uitverkoop''. Some English words like ``shop'' are even inflected in Dutch as ``shoppen'' and heavily used. While it is difficult in general to ascertain whether a foreign word or phrase used in an utterance is borrowed or just an instance of code-mixing~\cite{sharma2014borrowing}, one tell tale sign is that only proficient multilinguals can code-mix, while even monolingual speakers can use borrowed words because, by definition, these are part of the vocabulary of a language. In other words, just because an English speaker understands and uses the word ``tortilla'' does not imply that she can speak or understand Spanish. A borrowed word from $L_2$ initially appears frequently in speech, then gradually in print media like newspaper and finally it loses its origin's identity and is used in $L_1$ resulting in an inclusion in the dictionary of $L_1$ ~\cite{myers2002contact,thomason2003contact}. Borrowed words often take several years before they formally become part of $L_1$ dictionary. This motivates our research question ``is early-stage automatic identification of likely to be borrowed words possible?''. This is known to be a hard problem because (i) it is a socio-linguistic phenomenon closely related to acceptability and frequency, (ii) borrowing is a dynamic process; new borrowed words enter the lexicon of a language as old words, both native and borrowed, might slowly fade away from usage, and (iii) it is a population level phenomenon that necessitates data from a large portion of the population unlike standard natural language corpora that typically comes from a very small set of authors. Automatic identification of borrowed words in social media content (SMC) can improve language tagging by recommending the tagger to tag the language of the borrowed words as $L_1$ instead of $L_2$. The above reasons motivate us to resort to the social media (in particular, Twitter), where a large population of bilingual/multilingual speakers are known to often tweet in code-mixed colloquial languages ~\cite{carter2013microblog,solorio,vyas2014pos,JurgensLanguage2017,RijhwaniLanguageDetection2017}. We designed our methodology to work for any pair of languages $L_1$ and $L_2$ subject to the availability of sufficient SMC.  In the current study, we consider Hindi as $L_1$ and English as $L_2$.

The main stages of our research are as follows:

\noindent\textit{Metrics to quantify the likeliness of borrowing from social media signals}: We define three novel and closely similar metrics that serve as social signals indicating the likeliness of borrowing. We compare the likeliness of borrowing as predicted by our model and a baseline model with that from the ground truth obtained from human judges.

\noindent\textit{Ground truth generation}: We launch an extensive survey among $58$ human judges of various age groups and various educational backgrounds to collect responses indicating if each of the candidate foreign word is likely borrowed.


\noindent\textit{Application}: We randomly selected some words that have a high, low and medium borrowing likeliness as predicted by our metrics. Further, we randomly selected one tweet for each of the chosen words. The chosen words in almost all of these tweets have $L_2$ as their language tag while a majority of the surrounding words have a tag $L_1$. We asked expert annotators to re-evaluate the language tags of the chosen words and indicate if they would prefer to switch this tag from $L_2$ from $L_1$. 

Finally, our key results are outlined below:

\noindent 1. We obtained the Spearman's rank correlation between the ground-truth ranking and the ranking based on our metrics as $\sim 0.62$ for all the three variants which is more than double the value ($\sim 0.26$) if we use the most competitive baseline~\cite{sharma2014borrowing} available in the literature.

\noindent 2. Interestingly, the responses of the judges in the age group below $30$ seem to correspond even better with our metrics. Since language change is brought about mostly by the younger population, this might possibly mean that our metrics are able to capture the early signals of borrowing.

\noindent 3. Those users that mix languages the least in their tweets present the best signals of borrowing in case they do mix the languages (correlation of our metrics estimated from the tweets of these users with that of the ground truth is $\sim 0.65$).

\noindent 4. Finally, we obtain an excellent re-annotation accuracy of 88\% for the words falling in the surely borrowed category as predicted by our metrics.

%% file: 20Relatedwork.tex

\section{Related work}

In linguistics \textit{code-mixing} and \textit{borrowing} have been studied under the broader scope of language change and evolution. Linguists have for a long time focused on the sociological and the conversational necessity of borrowing and mixing in multilingual communities (see \citeauthor{auer} (\citeyear{auer}) and \citeauthor{muysken} (\citeyear{muysken}) for a review). In particular, \citeauthor{sankoff1990case} (\citeyear{sankoff1990case}) describes the complexity of choosing features that are indicative of borrowing. This work further showed that it is not always true that only highly frequent words are borrowed; nonce words could also be borrowed along with the frequent words. More recently, \cite{nzai2014understanding} analyzed the formal conversation of Spanish-English multilingual people and found that code mixing/borrowing is not only restricted to daily speech but is also prevalent in formal conversations.~\cite{hadei2016single} showed that phonological integration could be evaluated to understand the phenomenon of word borrowing. Along similar lines,~\cite{sebonde2014code} showed morphological and syntactic features could be good indicators for numerical borrowings.~\cite{senaratne2013borrowings} reported that in many languages English words are likely to be borrowed in both formal and semi-formal text.  

{\bf Mixing in computer mediated communication and social media}: ~\cite{sotillo} investigated various types of code-mixing in a corpora of 880 SMS text messages. The author observed that most often mixing takes place at the beginning of a sentence as well as through simple insertions. Similar observations about chat and email messages have been reported in~\cite{bock,negron}. However, studies of code-mixing with Chinese-English bilinguals from Hong Kong~\cite{li} and Macao~\cite{hksan} brings forth results that contrast the aforementioned findings and indicate that in these societies code-mixing is driven more by linguistic than social motivations.

Recently, the advent of social media has immensely propelled the research on code-mixing and borrowing as a dynamic social phenomena. ~\cite{hidayat} noted that in Facebook, users mostly preferred inter-sentential mixing and showed that 45\% of the mixing originated from real lexical needs, 40\% was used for conversations on a particular topic and the rest 5\% for content clarification. In contrast,~\cite{das} showed that in case of Facebook messages, intra-sentential mixing accounted for more than half of the cases while inter-sentential mixing accounted only for about one-third of the cases. In fact, in the First Workshop on \textit{Computational Approaches to Code Switching} a shared task on code-mixing in tweets was launched and four different code-mixed corpora were collected from Twitter as a part of the shared task~\cite{solorio}. Language identification task has also been handled for English-Hindi and English-Bengali code-mixed tweets in~\cite{das13}. Part-of-speech tagging have been recently done for code-mixed English-Hindi tweets~\cite{solorio2008part,vyas2014pos}.

{\bf Diachronic studies}: As an aside, it is interesting to note that the availability of huge volumes of timestamped data (tweet streams, digitized books) is now making it possible to study various linguistic phenomena quantitatively over different timescales. For instance,~\cite{sagi2009semantic} uses latent semantic analysis for detection and tracking of changes in word meaning, whereas ~\cite{frermann2016bayesian} presents a Bayesian approach for the same problem.~\cite{peirsman2010automatic} presents a distributed model for automatic identification of lexical variation between language varieties.~\cite{bamman2011measuring} discusses a method for automatically identifying word sense variation in a dated collection of historical books .~\cite{mitra2014s} presents a computational method for automatic identification of change in word senses across different timescales.~\cite{cook2014novel} presents a method for novel sense identification of words over time.

Despite these diverse and rich research agendas in the field of code-switching and lexical dynamics, there has not been much attempt to quantify the likeliness of borrowing of foreign words in a language. The only work that makes an attempt in this direction is~\cite{sharma2014borrowing}, which is described in detail in Sec 3.1.
One of the primary challenges faced by any quantitative research on lexical borrowing is that borrowing is a social phenomenon, and therefore, it is difficult to identify suitable indicators of such a lexical diffusion process unless one has access to a large population-level data. In this work, we show for the first time how certain simple and closely related signals encoding the language usage of social media users can help us construe appropriate metrics to quantify the likeliness of borrowing of a foreign word. 

%% file: 30Methodology.tex
\section{Methodology}
In this section, we present the baseline metric and propose three new metrics that quantify the likeliness of borrowing.

\subsection{Baseline metric}
\noindent{\em Baseline metric} -- We consider the $log(\frac{F_{L_2}}{F_{L_1}})$ value proposed in~\cite{sharma2014borrowing} as the baseline metric. Here $F_{L_2}$ denotes the frequency of the $L_1$ transliterated form of the word $w$ in the standard $L_1$ newspaper corpus. $F_{L_1}$, on the other hand, denotes the frequency of the $L_1$ translation of the word $w$ in the same newspaper corpus. For our experiments discussed in the later sections, both the transliteration and the translation of the words have been done by a set of volunteers who are native $L_1$ speakers. The authors in~\cite{sharma2014borrowing} claim that the more positive the value of this metric is for a word $w$, the higher is the likeliness of its being borrowed. The more negative the value is, the higher are the chances that the word $w$ is an instance of code-mixing. 

\noindent {\em Ranking} -- Based on the values obtained from the above metric for a set of target words, we rank these words; words with high positive values feature at the top of the rank list and words with high negative values feature at the bottom of the list. For two words having the same $log(\frac{F_{L_2}}{F_{L_1}})$ value, we resolve the conflict by assigning each of these the average of their two rank positions. In a subsequent section, we shall compare this rank list with the one obtained from the ground truth responses.

\subsection{Proposed metric}
In this section, we present three novel and closely related metrics based on the language usage patterns of the users of social media (in specific, Twitter). In order to define our metrics, we need all the words to be language tagged. The different tags that a word can have are: $L_1$, $L_2$, \textit{NE} (Named Entity) and \textit{Others}. Based on the word level tag, we also create a tweet level tag as follows: 
\begin{compactenum}
	\item $L_1$: Almost every word ($> 90\%$) in the tweet is tagged as $L_1$.
    \item $L_2$: Almost every word ($> 90\%$) in the tweet is tagged as $L_2$.
    \item $CML_1$: Code-mixed tweet but majority (i.e., $>50\%$) of the words are tagged as $L_1$.
	\item $CML_2$: Code-mixed tweet but majority (i.e., $>50\%$) of the words are tagged as $L_2$.
	\item \textit{CMEQ}: Code-mixed tweet having very similar number of words tagged as $L_1$ and $L_2$ respectively. 
	\item \textit{Code Switched}: There is a trail of $L_1$ words followed by a trail of $L_2$ words or vice versa.
\end{compactenum}  

Using the above classification, we define the following metrics:

\noindent{\em Unique User Ratio} ($UUR$) -- The Unique User Ratio for word usage across languages is defined as follows: 
\begin{equation}
UUR(w) = \frac{U_{L_1}+U_{CML_1}}{U_{L_2}}
\end{equation}
where $U_{L_1}$ ($U_{L_2}$, $U_{CML_1}$) is the number of unique users who have used the word $w$ in a $L_1$ ($L_2$, $CML_1$) tweet at least once.
 
\noindent{\em Unique Tweet Ratio} ($UTR$) -- The Unique Tweet Ratio for word usage across languages is defined as follows: 
\begin{equation}
UTR(w) = \frac{T_{L_1}+T_{CML_1}}{T_{L_2}}
\end{equation}
where $T_{L_1}$ ($T_{L_2}$, $T_{CML_1}$) is the total number of $L_1$ ($L_2$, $CML_1$) tweets which contain the word $w$.

\noindent{\em Unique Phrase Ratio} ($UPR$) -- The Unique Phrase Ratio for word usage across languages is defined as follows: 
\begin{equation}
UPR(w) = \frac{P_{L_1}}{P_{L_2}}
\end{equation}
where $P_{L_1}$ ($P_{L_2}$) is the number of $L_1$ ($L_2$) phrases which contain the word $w$. Note that unlike the definitions of $UUR$ and $UTR$ that exploit the word level language tags, the definition of $UPR$ exploits the phrase level language tags.

\noindent{\em Ranking} -- We prepare a separate rank list of the target words based on each of the three proposed metrics -- $UUR$, $UTR$ and $UPR$. The higher the value of each of this metric the higher is the likeliness of the word $w$ to be borrowed and higher up it is in the rank list. In a subsequent section, we shall compare these rank lists with the one prepared from the ground truth responses.


%% file: 40Experiments.tex
\section{Experiments}
In this section we discuss the dataset for our experiments, the evaluation criteria and the ground truth preparation scheme.

\subsection{Datasets and preprocessing}
In this study, we consider code-mixed tweets gathered from Hindi-English bilingual Twitter users in order to study the effectiveness of our proposed metrics. The native language $L_1$ is Hindi and the foreign language $L_2$ is English. To bootstrap the data collection process, we used the language tagged tweets presented in~\cite{rudra2016understanding}. In addition to this, we also crawled tweets (between Nov 2015 and Jan 2016) related to 28 hashtags representing different Indian contexts covering important topics such as sports, religion, movies, politics etc. This process results in a set of 811981 tweets. We language-tag (see details later in this section) each tweet so crawled and find that there are 3577 users who use mixed language for tweeting. Next, we systematically crawled the time lines of these 3577 users between Feb 2016 and March 2016 to gather more mixed language tweets. Using this two step process we collected a total of 1550714 distinct tweets. From this data, we filtered out tweets that are not written in romanized script, tweets having only URLs and tweets having empty content. Post filtering we obtained 725173, tweets which we use for the rest of the analysis. The datasets can be downloaded from \url{http://cnerg.org/borrow}

\noindent{\bf Language tagging}: We tag each word in a tweet with the language of its origin using the method outlined in~\cite{gella2013query}. \textit{Hi} represents a predominantly Hindi tweet, \textit{En} represents a predominantly English tweet, \textit{CMH} (\textit{CME}) represents code-mixed tweets with more Hindi (English) words, \textit{CMEQ} represents code-mixed tweets with almost equal number of Hindi and English words and \textit{CS} represents code-switched tweets (the number and \% of tweets in each of the above six categories are presented in the supplementary material). Like the word level, the tagger also provides a phrase level language tag. Once again, the different tags that an entire phrase can have are: \textit{En}, \textit{Hi} and \textit{Oth} (Other). The metrics defined in the previous section are computed using these language tags.

\noindent{\bf Newspaper dataset for the baseline}: As we had discussed in the previous section, for the construction of the baseline ranking we need to resort to counting the frequency of the foreign words (i.e., English words) and their Hindi translations in a newspaper corpus as has been outlined in~\cite{sharma2014borrowing}. For this purpose, we use the FIRE dataset built from the Hindi Jagaran newspaper corpus\footnote{Jagaran corpus: \url{http:/fire.irsi.res.in/fire/static/data}} which is written in Devanagari script.

\subsection{Target word selection}

We first compute the most frequent foreign (i.e., English words) in our tweet corpus. Since we are interested in the frequency of the English word only when it appears as a foreign word we do not consider the (i) \textit{Hi} tweets since they do not have any foreign word, (ii) \textit{En} tweets since here the English words are not foreign words and the (iii) code-switched tweets. Based on the frequency of usage of English as a foreign word, we select the top 1000 English words. Removal of stop words and text normalization leaves beyond 230 nouns (see supplementary material for the list of words). 

\noindent{\bf Final selection of target words}: In language processing, context plays an important role in understanding different properties of a word. For our study, we also attempt to use the language tags as features of the context words for a given \TG. Our hypothesis here is that there should exist classes of words that have similar context features and the likelihood of being borrowed in each class should be different. For example, when an English word is surrounded by mostly Hindi words it seems to be more likely borrowed. We present two examples in the box below to illustrate this. 
\vspace{2mm} 
\fbox{\begin{centering}\begin{minipage}{0.43\textwidth}
{\em Example I}:\\
@****** Welcome. \textbf{\textit{Film}} \textit{jaroor dekhna. Nahi to injection ready hai.} \\ 
{\em Example II}:\\
@*****\_Trust @*****\\
\textit{Kuch to ache se karo sirji}....\\
\textit{Har jagah bhaagte rehna} is not a good \textbf{\textit{thing}}.\\
\end{minipage}
\end{centering}
}
In \textit{Example I} the English word ``film'' is surrounded by mostly Hindi words. On the other hand, in \textit{Example II} the English word ``thing'' is surrounded mostly by English words. Note that the word ``film'' is very commonly used by Hindi monolingual speakers and is therefore highly likely to have been borrowed unlike the English word ``thing'' which is arguably an instance of mixing. This socio-linguistic difference seems to be very appropriately captured by the language tag of the surrounding words of these two words in the respective tweets. Based on this hypothesis we arrange the 230 words into contextually similar groups (see supplementary material for the grouping details). Finally, using the baseline metric $log(\frac{F_E}{F_H})$ ($E$: English, $H$: Hindi), we proportionately choose words from these groups as follows:\\
\noindent{\em Words with very high or very low values of} $log(\frac{F_E}{F_H})$ ($hlws$) -- we select words having the highest and the lowest values of $log(\frac{F_E}{F_H})$ from each of the context groups. This constitutes a set of 30 words. Note that these words are baseline-biased and therefore the metric should be able to discriminate them well.\\
\noindent{\em Words with medium values of} $log(\frac{F_E}{F_H})$ ($mws$) -- we selected 27 words having not so high and not so low $log(\frac{F_E}{F_H})$ at uniformly at random.\\
\noindent{\em Full set of words} ($full$) -- Thus, in total we selected 57 target words for the purpose of our evaluation. We present these words in the box below.

\fbox{\begin{minipage}{0.43\textwidth}\small
\textit{Baseline-biased words} -- \textit{'thing', 'way', 'woman', 'press', 'wrong', 'well', 'matter', 'reason', 'question', 'guy', 'moment', 'week', 'luck', 'president', 'body', 'job', 'car', 'god', 'gift', 'status', 'university', 'lyrics', 'road', 'politics', 'parliament', 'review', 'scene', 'seat', 'film', 'degree'}\\
\textit{Randomly selected words} -- \textit{ 'people', 'play', 'house', 'service', 'rest', 'boy', 'month', 'money', 'cool', 'development', 'group', 'friend', 'day', 'performance', 'school', 'blue', 'room', 'interview', 'share', 'request', 'traffic', 'college', 'star', 'class', 'superstar', 'petrol', 'uncle'}
\end{minipage}}

\subsection{Evaluation criteria}
We present a four step approach for evaluation as follows. We measure (i) how well the $UUR$, $UTR$ and $UPR$ based ranking of the $hlws$ set, the $mws$ set and the $full$ set correlate with the ground truth ranking (discussed in the next section) in comparison to the rank given by the baseline metric, (ii) how well the different rank ranges obtained from our metric align with the ground truth as compared to the baseline metric, (iii) whether there are some systematic effects of the age group of the survey participants on the rank correspondence,  (iv) how our metrics if computed from the tweets of users who (a) rarely mix languages, (b) almost always mix languages and (c) are in between (a) and (b), align with the ground truth.

\noindent{\bf Rank correlation}: We measure the standard Spearman's rank correlation ($\rho$)~\cite{zar1972significance} pairwise between rank lists generated by (i) $UUR$ (ii) $UTR$ (iii) $UPR$ (iv) baseline metric and the ground truth.  

We shall describe the next four measurements taking $UUR$ as the running example. The same can be extended verbatim for the other two similar metrics.

\noindent{\bf Rank ranges}: We split each of the three rank lists ($UUR$, ground truth and baseline) into five different equal-sized ranges as follows -- (i) surely borrowed (SB) containing top 20\% words from each list, (ii) likely borrowed (LB) containing the next 20\% words from each list, (iii) borderline (BL) constituting the subsequent 20\% words from each list, (iv) likely mixed (LM) comprising the next 20\% words from each list and (v) surely mixed (SM) having the last 20\% words from each rank list. Therefore, we have three sets of five buckets, one set each for $UUR$, the ground truth and the baseline based rank list. 

Next we calculate the bucket-wise correspondence between (i) the $UUR$ and the ground truth set and (ii) the baseline and the ground truth set in terms of standard \textit{precision} and \textit{recall} measures. For our purpose, we adapt these measures as follows.\\
$G$: ground truth bucket set, $B_b$: baseline bucket set, $U_b$: $UUR$ bucket set;\\
$BS \in \{B_b, U_b\}$, $T$ (type of bucket) = \{SB, LB, BL, LM, SM\};\\
$b_t$ = words in type $t$ bucket from $BS$, $g_{t}$ = words in type $t$ bucket from $G$, $t \in T$;\\ 
$tp_{t}$ (no. of true positives) = $|b_{t} \cap g_{t}|$, $fp_{t}$ (no. of false positives) = $|b_{t} - g_{t}|$, $tn_{t}$ (no. of true negatives) = $|g_{t} - b_{t}|$; \\ 
Bucket-wise \textit{precision} and \textit{recall} therefore:\\ $precision(b_{t}) = \frac{tp_{t}}{fp_{t} + tp_{t}}$; $recall(b_{t}) = \frac{tp_{t}}{tn_{t} + tp_{t}}$\\
For a given set, we obtained the overall \textit{macro precision} (\textit{recall}) by averaging the \textit{precision} (\textit{recall}) values over the five buckets. For a given set, we also obtained the overall \textit{micro precision} by first adding the true positives across all the buckets and then normalizing by the sum of the true and the false positives over all the buckets. We take an equivalent approach for obtaining the \textit{micro recall}.

\noindent{\bf Age group effect}: Here we construct two ground truth rank lists one using the responses of the participants with age below 30 (young population) and the other using the responses of the rest of the participants (elderly population). Next we repeat the above two evaluations considering each of the new ground truth rank lists. 

\noindent{\bf Extent of language mixing}: Here we divide all the 3577 users into three categories -- (i) High (users who have more than 20\% of tweets as code-mixed), (ii) Mid (users who have 7--20\% of their tweets as code-mixed, and (iii) Low (users who have less than 7\% of their tweets as code-mixed). We create three $UUR$ based rank lists for each of these three user categories and respectively compare them with the ground truth rank list.

\subsection{Ground truth preparation}

Since it is very difficult to obtain a suitable ground truth to validate the effectiveness of our proposed ranking scheme, we launched an online survey to collect human judgment for each of the 57 \TG\textit{s}. 

\noindent{\bf Online survey}
We conducted the online survey\footnote{Survey portal: \url{https://goo.gl/forms/L0kJm8BNMhRj0jA53}} among 58 volunteers majority of whom were either native language(Hindi) speakers or had very high proficiency in reading and writing in that language. The participants were selected from different age groups and different educational backgrounds. Every participant was asked to respond to a multiple choice question about each of the 57 \TG\textit{s}. Therefore, for every single \TG, 58 responses were gathered. The multiple choice question had the following three options and the participants were asked to select the one they preferred the most and found more natural -- (i) a Hindi sentence with the \TG~as the only English word, (ii) the same Hindi sentence in (i) but with the \TG~replaced by its Hindi translation and (iii) none of the above two options. There were no time restrictions imposed while gathering the responses, i.e., the volunteers theoretically had unlimited time to decide their responses for each \TG. 

\noindent{\bf Language preference factor}
For each \TG, we compute a \textit{language preference factor} ($LPF$) defined as $(Count_{En}-Count_{Hi})$, where $Count_{Hi}$ refers to the number of survey participants who preferred the sentence containing the Hindi translation of the \TG~while $Count_{En}$ refers to the number of survey participants who preferred the sentence containing the \TG~itself. More positive values of $LPF$ denotes higher usage of \TG~as compared to its Hindi translation and therefore higher likeliness of the word being borrowed.

\noindent{\bf Ground truth rank list generation}
We generate the ground truth rank list based on the $LPF$ score of a \TG. The word with the highest value of $LPF$ appears at the top of the ground truth rank list and so on in that order. Tie breaking between \TG\textit{s} having equal $LPF$ values is done by assigning average rank to each of these words.

\noindent{\bf Age group based rank list}: As discussed in the previous section, we prepare the age group based rank lists by first splitting the responses of the survey participants in two groups based on their age -- (i) young population (age $< 30$) and (ii) elderly population (age $\ge 30$). For each group we then construct a separate $LPF$ based ranking of the \TG\textit{s}. 

%% file: 50Results.tex
\section{Results}\label{sec:eval}

\subsection{Correlation among rank lists}
The Spearman's rank correlation coefficient ($\rho$) of the rank lists for the $hlws$ set, the $mws$ set and the $full$ set according to the baseline metric, $UUR$, $UTR$ and $UPR$ with respect to the ground truth metric $LPF$ are noted in table~\ref{tab:SpearmanValue}. We observe that for the $full$ set, the $\rho$ between the rank lists obtained from all the three metrics $UUR$, $UTR$, and $UPR$ with respect to the ground truth is $\sim 0.62$ which is more than double the $\rho$ ($\sim 0.26$) between the baseline and the ground truth rank list. This clearly shows that the proposed metrics are able to identify the likeliness of borrowing quite accurately and far better than the baseline. Further, a remarkable observation is that our metrics outperform the baseline metric even for the $hlws$ set that is baseline-biased. Likewise, for the $mws$ set, our metrics outperform the baseline indicating a superior recall on arbitrary words. The ranking of the $full$ set of words obtained from the ground truth, the baseline and the $UUR$ metric is available in the supplementary material. 


We present the subsequent results for the $full$ set and the $UUR$ metric. The results obtained for the other two metrics $UTR$ and $UPR$ are very similar and therefore not shown.

\begin{table}
\centering
\scalebox{0.65}{ 
\begin{tabular}{|l|l|l|l|l|} 
\hline
Rank-List$_1$ & Rank-List$_2$ & $\rho-hlws$ & $\rho-mws$ & $\rho-full$ \\ \hline \hline
$UUR$         & Ground truth     & \textbf{0.67}    & \textbf{0.64}      & \textbf{0.62}             \\ \hline
$UTR$         & Ground truth	  &\textbf{0.66}      & \textbf{0.63}     & \textbf{0.63}             \\ \hline
$UPR$         & Ground truth      &\textbf{0.66}       &\textbf{0.64}    & \textbf{0.62}             \\ \hline
Baseline   & Ground truth      &0.49     &0.14        & 0.26                         \\ \hline
\end{tabular}
}
\caption{Spearman's rank correlation coefficient ($\rho$) among the different rank lists. Best result is marked in bold.}
\label{tab:SpearmanValue}
\end{table}

\subsection{Rank list alignment across rank ranges}
The number of \TG\textit{s} falling in each bucket across the three rank lists are the same and are noted in table~\ref{tab:BucketWordDistribution}. Thus, the \textit{precision} and \textit{recall} as per the definition are also the same. The bucket-wise \textit{precision}/\textit{recall} for the baseline and $UUR$ with respect to the ground truth are noted in table~\ref{tab:BucketWisePR}. We observe that while in the SB bucket both the baseline and $UUR$ perform equally well, for all the other buckets $UUR$ massively outperforms the baseline. This implies that for the case where the likeliness of borrowing is the strongest, the baseline does as good as $UUR$. However, as one moves down the rank list, $UUR$ turns out to be a considerably better predictor than the baseline. The overall \textit{macro} and \textit{micro} \textit{precision}/\textit{recall} as shown in table~\ref{tab:OverallMiMaPrRc} further strengthens our observation that $UUR$ is a better metric than the baseline.

\begin{table}
\centering
\scalebox{0.5}{ 
\begin{tabular}{|c|c|c|c|}
\hline
Bucket type & Ground truth bucket   & Baseline bucket  & $UUR$ bucket 		\\ \hline \hline
SB         & 11      	& 11 			& 11	\\ \hline
LB  	 	& 11      	& 11 			& 11    \\ \hline
BL         & 12      	& 12 			& 12    \\ \hline
LM        	& 11      	& 11 			& 11    \\ \hline
SM         	& 12      	& 12 			& 12    \\ \hline
\end{tabular} }
\caption{Number of words falling in each bucket of three bucket sets.}
\label{tab:BucketWordDistribution}
\end{table}

\begin{table}
\centering
\scalebox{0.6}{
\begin{tabular}{|c|c|c|}
\hline
Bucket type & \textit{prec.}/\textit{rec.} Baseline   & \textit{prec.}/\textit{rec.} $UUR$  		\\ \hline \hline
SB         & \textbf{0.27}     	& \textbf{0.27} 			                \\ \hline
LB  	 	& 0.09      	& \textbf{0.18}			                   \\ \hline
BL         & 0.08   		& \textbf{0.33} 			                   \\ \hline
LM        	& 0.18   		& \textbf{0.36} 			                   \\ \hline
SM         	& 0.33   		& \textbf{0.50} 			                   \\ \hline
\end{tabular}}
\caption{Bucket-wise \textit{precision}/\textit{recall}. Best results are marked in bold.}
\label{tab:BucketWisePR}
\end{table}

\begin{table}
\vspace*{-4mm}
\centering
\scalebox{0.6}{
\begin{tabular}{|c|c|c|}
\hline
Measure &  Baseline   &  $UUR$  		\\ \hline \hline
\textit{Macro prec.}/\textit{rec.}         & 0.19      	& \textbf{0.33} 			                \\ \hline
\textit{Micro prec.}/\textit{rec.}         & 0.19   			& \textbf{0.33}  			            \\ \hline
\end{tabular} }
\caption{  Overall \textit{macro} and \textit{micro} \textit{precision}/\textit{recall}. Best results are marked in bold.}
\label{tab:OverallMiMaPrRc}
\end{table}

\subsection{Age group based analysis}
As already discussed earlier, we split the ground truth responses based on the age group of the survey participants. We split the responses into two groups -- (i) young population (age $< 30$) and (ii) elderly population (age $\ge 30$) so that there are almost equal number of responses in both the groups (see supplementary material for the exact distribution).

\noindent{\bf Rank correlation}: The Spearman's rank correlation of $UUR$ and the baseline rank lists with these two ground truth rank lists are shown in table~\ref{tab:SpearManAcrossAge}. Interestingly, the correlation between $UUR$ rank list and the young population ground truth is better than the elderly population ground truth. This possibly indicates that $UUR$ is able to predict recent borrowings more accurately. However, note that the $UUR$ rank list has a much higher correlation with both the ground truth rank lists as compared to the baseline rank list.

\begin{table}
\centering
\scalebox{0.6}{ 
\begin{tabular}{|l|l|l|}
\hline
Rank-List$_1$ & Rank-List$_2$ & $\rho$  \\ \hline \hline
Baseline   & Ground-truth-Young      & 0.26 \\ \hline
$UUR$         & Ground-truth-Young      & \textbf{0.62}  \\ \hline
Baseline   & Ground-truth-Elder      & 0.27 \\ \hline
$UUR$         & Ground-truth-Elder      & \textbf{0.53} \\ \hline
\end{tabular}
}
\caption{Spearman's rank correlation across the two age groups. Best results are marked in bold.}
\label{tab:SpearManAcrossAge}
\end{table}

\noindent{\bf Rank ranges}: Table ~\ref{tab:NumWrdsYagOagGT} shows the bucket-wise \textit{precision} and \textit{recall} for $UUR$ and the baseline metrics with respect to two new ground truths. For the young population once again the number of words in each bucket for all the three sets is the same thus making the values of the \textit{precision} and the \textit{recall} same. In fact, the \textit{precision}/\textit{recall} for this ground truth is exactly same as in the case of the original ground truth. 

In contrast, when we consider the ground truth based on the responses of the elderly population, the number of words across the different buckets are different across the three sets. In this case, we observe that the \textit{precision}/\textit{recall} values are better for the $UUR$ metric in SB, LB and LM buckets. 

\begin{table}
\centering
 \scalebox{0.65}{ 
\begin{tabularx}{0.58\textwidth}{| X | X | X | X | X | X | X |} 
\hline
Bucket type & Young-Baseline \textit{p}/\textit{r} & Young-$UUR$ \textit{p}/\textit{r}  & Elder-Baseline \textit{p} & Elder-$UUR$ \textit{p} & Elder-baseline \textit{r} & Elder-$UUR$ \textit{r} 		\\ \hline \hline
SB         &\textbf{0.27}     &\textbf{0.27}  	&0.27     &\textbf{0.36}  	&0.25		&\textbf{0.33} \\ \hline
LB  	 	&0.09     &\textbf{0.18}  	&0.09     &\textbf{0.18}  	&0.08		&\textbf{0.17}	\\ \hline
BL         &0.08     &\textbf{0.33}		&\textbf{0.16}     &0.08  	&\textbf{0.28}		&0.14	\\ \hline
LM        	&0.18     &\textbf{0.36}		&0.18     &\textbf{0.45}  	&0.14		&\textbf{0.35}	\\ \hline
SM         	&0.33     &\textbf{0.5}   		&\textbf{0.41}     &0.25    	&\textbf{0.41}		&0.25  	\\ \hline
 
\end{tabularx} 
} 
\caption{Bucket-wise \textit{precision} (\textit{p})/\textit{recall} (\textit{r}) for $UUR$ and the $baseline$ metrics for the two new ground truths. Best results are marked in bold.}
\label{tab:NumWrdsYagOagGT}
\end{table} 

Finally, the overall  \textit{macro} and \textit{micro} \textit{precision} and \textit{recall} for both the age groups are noted in table~\ref{tab:OverallMiMaAge}. Once again, for both the young and the elderly population based ground truths, the \textit{macro} and \textit{micro precision} and \textit{recall} values for the $UUR$ metric are higher compared to that of the baseline. 

\begin{table}
\centering
\scalebox{0.65}{ 
\begin{tabularx}{0.6\textwidth}{|X|X|X|X|X|}
\hline
Measure 					&  Young-Baseline    	&  Young-$UUR$ 		&  Elder-Baseline    	& Elder-$UUR$	 	\\ \hline \hline
\textit{Mac.} \textit{pre.} 	&  0.19			    &  \textbf{0.33} 			&  0.22            	&  \textbf{0.27}			\\ \hline
\textit{Mac.} \textit{rec.}  		&  0.19     			&  \textbf{0.33}			&  0.23				&  \textbf{0.25}			\\ \hline
\textit{Mic. pre.}/\textit{rec.} 	&  0.19			&  \textbf{0.33} 			&  0.23	  	 	& \textbf{0.26}         	\\ \hline

\end{tabularx}
}
\caption{Overall \textit{macro} and \textit{micro} \textit{precision} and \textit{recall} for the two new ground truths. Best results are marked in bold.}
\label{tab:OverallMiMaAge}
\end{table}

\subsection{Extent of language mixing}

As mentioned earlier, we divide the set of 3577 users into three categories. The Spearman's correlation between $UUR$ and the ground truth for each of these buckets are given in table~\ref{tab:UserSetBucket}. As we can see, for Low bucket the $\rho$ value is maximum. This points to the fact that the signals of borrowing is strongest from the users who rarely mix languages.

\begin{table}
\centering
\scalebox{0.6}{ 
\begin{tabularx}{0.6\textwidth}{|X|X|X|}
\hline
Bucket &  Number of users &    $\rho$				 	\\ \hline \hline
High   &  	302          & 	 0.52		\\ \hline
Mid   &  	744          & 	 0.60		\\ \hline
Low   &  	2531         & 	\textbf{0.65}		\\ \hline 

\end{tabularx}
}
\caption{Spearman's correlation between $UUR$ and the ground truth in the different user buckets. Best results are marked in bold.}
\label{tab:UserSetBucket}
\end{table}

\section{Re-annotation results}

In order to conduct the re-annotation experiments we performed the following. To begin with, we ranked all the 230 English nouns in non-increasing order of their $UUR$ values. We then randomly selected 20 words each having (i) high $UUR$ (top 20\%) values (call $TOP$), (ii) low $UUR$ (bottom 20\%) values (call $BOT$), and (iii) middle $UUR$ (middle 20\%) values (call $MID$). This makes a total of 60 words. Using this word list we extracted one tweet each that contained the (foreign) word from this list along with all other words in the tweet tagged in Hindi ($H_{all}$). We similarly prepared another such list of 60 words and extracted one tweet each in which most of the other words were tagged in Hindi ($H_{most}$). 

We presented the selected words and the corresponding tweets to a set of annotators and asked them to annotate these selected words once again. Over all the words, we calculated the mean ($\mu_{E\rightarrow H}$) and the standard deviation ($\sigma_{E\rightarrow H}$) of the fraction of cases where the annotators altered the tag of the selected word from English to Hindi. The average inter-annotator agreement~\cite{fleiss1971measuring} for our experiments was found to be as high as 0.64. For the words in the $TOP$ list, the fraction of times the tag is altered is \textbf{0.91} (\textbf{0.85}) with an inter-annotator agreement of 0.84 (0.80) for the $H_{all}$ ($H_{most}$) category. In other words, on average, in as high as 88\% of cases the annotators altered the tags of the words that are highly likely to be borrowed (i.e., $TOP$) in a largely Hindi context (i.e., $H_{all}$ or $H_{most}$). Table~\ref{tab:AnotationResults} shows the fractional changes for all the other possible cases. An interesting observation is that the annotators rarely flipped the tags for the words in the $BOT$ list (i.e., the sure mixing cases) in either of the $H_{all}$ or the $H_{most}$ contexts. These results strongly support the inclusion of our metric in the design of future automatic language tagging tasks. 

\begin{table}
\centering
\scalebox{0.6}{ 
\begin{tabularx}{0.6\textwidth}{|X|X|X|X|}
\hline
Word-rank &  Context & $\mu_{E\rightarrow H}$ & $\sigma_{E\rightarrow H}$  \\ \hline \hline
$TOP$   &  	$H_{all}$  &   \textbf{0.91}     &     0.15		\\ \hline
$TOP$   &  	$H_{most}$ &   \textbf{0.85}     &  	 0.23		\\ \hline
$MID$   &  	$H_{all}$ &    0.58     &  	 0.28		\\ \hline
$MID$   &  	$H_{most}$ &   0.61     &  	 0.34		\\ \hline
$BOT$   &  	$H_{all}$ &    0.13     &  	 0.18		\\ \hline
$BOT$   &  	$H_{most}$ &   0.16     &  	 0.21		\\ \hline

\end{tabularx}
}
\caption{Re-annotation results}
\label{tab:AnotationResults}
\end{table}

%% file: 60Conclusion.tex
\section{Discussion and conclusion}
In this paper, we proposed a few new metrics for estimating the likliness of borrowing that rely on signals from large scale social media data. Our best metric is two-fold better than the previous metric in terms of the accuracy of the prediction. There are some interesting linguistic aspects of borrowing as well as certain assumptions regarding the social media users that have important repercussions on this work and its potential extensions,  which are discussed in this section.

\textbf{Types of borrowing}: Linguists define broadly three forms of borrowing, (i) \textit{cultural}, (ii) \textit{core}, and (iii) \textit{therapeutic} borrowings. In \textit{cultural borrowing}, a foreign word gets borrowed into native language to fill a lexical gap. This is because there is no equivalent native language word present to represent the same foreign word concept. For instance, the English word `computer' has been borrowed in many Indian languages since it does not have a corresponding term in those languages\footnote{Words for ``computer" were coined in many Indian languages through morphological derivation from the term for ``compute'', however, none of these words are used in either formal or informal contexts.}. In {\em core borrowing}, on the other hand, a foreign word replaces its native language translation in the native language vocabulary. This occurs due to overwhelming use of the foreign word over native language translation as a matter of prestige, ease of use etc. For example, the English word `school' has become much more prevalent than its corresponding Hindi translation `{\em vidyalaya}' among the native Hindi speakers. Finally, therapeutic borrowing refers to borrowing of  words to avoid taboo and homonomy in the native language. In this paper, although we did not perform any category based studies, most of our focus was on core borrowing.

\textbf{Language of social media users}: We assumed that if a user is predominantly using Hindi words in a tweet then the chances of him/her being a native Hindi speaker should be high, since, while the number of English native speakers in India is 0.02\%, the number of Hindi native speakers is 41.03\%\footnote{http://bit.ly/2ufOvea}. This assumption has also been made in earlier studies~\cite{rudra2016understanding}. Note that even if a user is not a native Hindi speaker but a proficient (or semi-proficient) Hindi speaker, the main results of our analysis should hold. For instance, consider two foreign words `a' and `b'. If `a' is frequently borrowed in the native language, then the proficient speaker would also tend to borrow `a' similar to a native speaker. Even if due to lack of adequate native vocabulary, the non-native speaker borrows the word `b' in some cases, these spurious signals should get eliminated since we are making an aggregate level statistics over a large population.

\textbf{Future directions}: It would be interesting to understand and develop theoretical justification for the metrics. Further, it would be useful to study and classify various other linguistic phenomena closely related to core borrowing, such as: (i) \textit{loanword}, where a form of a foreign word and its meaning or one component of its meaning gets borrowed, (ii) \textit{calques}, where a foreign word or idiom is translated into existing words of native language, and (iii) \textit{semantic loan}, where the word in the native language already exists but an additional meaning is borrowed from another language and added to existing meaning of the word.

Finally, we would also like to incorporate our findings into other standard tasks of multilingual IR and multilingual speech synthesis (for example to render the appropriate native accent to the borrowed word).